\documentclass{article} 
\usepackage{iclr2024_conference,times}

\usepackage{amsmath,amsfonts,bm}

\def\eqref#1{equation~\ref{#1}}

\def\1{\bm{1}}

\DeclareMathAlphabet{\mathsfit}{\encodingdefault}{\sfdefault}{m}{sl}
\SetMathAlphabet{\mathsfit}{bold}{\encodingdefault}{\sfdefault}{bx}{n}

\usepackage{url}

\usepackage{setspace}
\usepackage{url}
\usepackage{booktabs}       
\usepackage{amsfonts}       
\usepackage{amsmath}
\usepackage{amsthm}
\usepackage{amssymb}
\usepackage{nicefrac}       
\usepackage{microtype}      
\usepackage{xcolor}         
\usepackage{graphicx}

\usepackage[normalem]{ulem}
\usepackage{algorithm}
\usepackage[noend]{algpseudocode}
\usepackage{xspace}
\usepackage{float}
\usepackage{tikz}
\usepackage{verbatim}
\usepackage{tabularx}
\usepackage{enumitem}
\usepackage{wrapfig}
\usepackage{multicol}
\usepackage[most]{tcolorbox}
\usepackage[framemethod=tikz]{mdframed}
\usepackage{subcaption}
\usepackage{caption}
\usepackage{multirow}

\usepackage{pifont}
\newcommand{\cmark}{\text{\ding{51}}}
\newcommand{\xmark}{\text{\ding{55}}}

\usepackage{listings}
\usepackage{color}
\usepackage{xcolor}

\usepackage[draft=true]{minted}

\usepackage{float}
\usepackage{graphicx}

\definecolor{isarblue}{HTML}{006699}
\definecolor{isarfaintblue}{rgb}{0.0, 0.75, 1.0}
\definecolor{isargreen}{HTML}{009966}
\definecolor{red}{HTML}{990000}
\definecolor{patriarch}{rgb}{0.5, 0.0, 0.5}

\definecolor{citecolor}{HTML}{2980b9}
\definecolor{linkcolor}{HTML}{c0392b}
\usepackage[pagebackref=true,breaklinks=true,colorlinks,bookmarks=false,citecolor=citecolor,linkcolor=linkcolor]{hyperref}

\lstdefinelanguage{isabelle}{%
    keywords=[1]{type_synonym,datatype,fun,abbreviation,definition,proof,lemma,theorem,qed,corollary,have,hence,also,finally,ultimately,moreover,using,\{},
    keywordstyle=[1]\bfseries\color{isarblue},
    keywords=[2]{where,assumes,shows,fixes,and},
    keywordstyle=[2]\bfseries\color{isargreen},
    keywords=[3]{if,then,else,case,SOME,let,in,O},
    keywordstyle=[3]\color{isarblue},
    keywords=[4]{ATPWithTC},
    keywordstyle=[4]\it\color{patriarch},
    keywords=[5]{show,assume,obtain},
    keywordstyle=[5]\bfseries\color{isarfaintblue},
}

\lstdefinestyle{isabelle}{%
  language=isabelle,
  escapeinside={\&}{&},
  columns=fixed,
  extendedchars,
  basewidth={0.5em,0.45em},
  basicstyle=\singlespacing\ttfamily\tiny,
  mathescape,
  morecomment=[s][\bfseries\color{red}]{(*}{*)},
  morecomment=[l][\bfseries]{####},
}

\definecolor{mybrown}{RGB}{128,64,0}
\gdef\Sepline{%
  \par\noindent\makebox[\linewidth][l]{%
  \hspace*{-\mdflength{innerleftmargin}}%
   \tikz\draw[thick,dashed,gray!60] (0,0) --%
        (\textwidth+\the\mdflength{innerleftmargin}+\the\mdflength{innerrightmargin},0);
  }\par\nobreak}

\def\miniff{miniF2F\xspace}

\def\methodOne{TC\xspace}
\def\methodOneFull{\textit{Tool Correction}\xspace}
\def\methodTwo{CC\xspace}
\def\methodTwoFull{\textit{Conjecture Correction}\xspace}
\def\fullname{Lyra\xspace}
\def\shortname{Lyra\xspace}

\title{Lyra: Orchestrating Dual Correction in Automated Theorem Proving}

\author{%
  Chuanyang Zheng$^{1}$,\space\space\space
  Haiming Wang$^{2}$,\space\space\space
  Enze Xie$^{3}$,\space\space\space
  Zhengying Liu$^{3}$,\space\space\space
  \\ \vspace{0.15cm}
  \textbf{Jiankai Sun$^{1}$,\space\space\space
  Huajian Xin$^{2}$,\space\space\space
  Jianhao Shen$^{3}$,\space\space\space
  Zhenguo Li$^{3}$,\space\space\space
  Yu Li$^{1}$\space\space\space
  }
  \\
  $^{1}$\normalfont{The Chinese University of Hong Kong}\quad
  $^{2}$\normalfont{Sun Yat-sen University}\quad\\
  $^{3}$\normalfont{Huawei Noah’s Ark Lab}\quad
  \vspace{0.15cm} \\
  {\tt\small \{cyzheng21,liyu\}@cse.cuhk.edu.hk,
  \{wanghm39,xinhj\}@mail2.sysu.edu.cn}\\
  {\tt\small\{xie.enze, liuzhengying2, shenjianhao2, Li.Zhenguo\}@huawei.com}\\
  {\tt\small\{jsun\}@link.cuhk.edu.hk}
  \AND
  \quad\quad\space\space\space\url{https://github.com/chuanyang-Zheng/Lyra-theorem-prover}
}

\iclrfinalcopy 
\begin{document}

\maketitle

\begin{abstract}
Large Language Models (LLMs) present an intriguing avenue for exploration in the field of formal theorem proving. Nevertheless, their full potential, particularly concerning the mitigation of hallucinations and refinement through prover error messages, remains an area that has yet to be thoroughly investigated. To enhance the effectiveness of LLMs in the field, we introduce the \fullname, a new framework that employs two distinct correction mechanisms: \methodOneFull (\methodOne) and \methodTwoFull (\methodTwo).
To implement \methodOneFull in the post-processing of formal proofs, we leverage prior knowledge to utilize predefined prover tools (e.g., Sledgehammer) for guiding the replacement of incorrect tools. \methodOneFull significantly contributes to mitigating hallucinations, thereby improving the overall accuracy of the proof. 
In addition, we introduce \methodTwoFull, an error feedback mechanism designed to interact with prover to refine formal proof conjectures with prover error messages. Compared to the previous refinement framework, the proposed \methodTwoFull refines generation with instruction but does not collect paired (generation, error \& refinement) prompts.
Our method has
achieved state-of-the-art (SOTA) performance on both miniF2F validation ($48.0\%$ $\rightarrow$ $55.3\%$) and test ($45.5\%$ $\rightarrow$ $51.2\%$). 
We also present 3 IMO problems solved by \fullname.
We believe \methodOneFull (post-process for hallucination mitigation) and \methodTwoFull (subgoal adjustment from interaction with environment) could provide a promising avenue for future research in this field.
\end{abstract}

\section{Introduction}
Formal proof automation is a challenging task that has garnered increased attention in recent years~\citep{bansal2019holist,polu2020generative,lample2022hypertree,jiang2022thor,wu2022autoformalization, wang2023dt}. Unlike other domains where deep learning approaches have shown remarkable success, previous studies have proposed techniques to synthesize additional formal training data~\citep{wu2022autoformalization,polu2020generative,han2021proof, bansal2019learning,polu2023formal}. Recently, large language models (LLMs) trained on informal mathematical data have showcased impressive quantitative reasoning abilities~\citep{lewkowycz2022solving, welleck2022naturalprover}.

Draft, Sketch, and Prove (DSP)~\citep{jiang2023draft} maps informal proofs to formal proof sketches, and uses the sketches to guide an automated prover by directing its search to easier sub-problems. Following this direction, Subgoal-based Learning~\citep{zhao2023decomposing} replaces the informal proof with subgoal-proof and learns how to optimize subgoal demonstration selection. However, they have not been able to post-process LLM generation or gradually refine previous generations.

\begin{figure}[ht]
\setlength{\abovecaptionskip}{0.1cm}
\includegraphics[width=\textwidth]{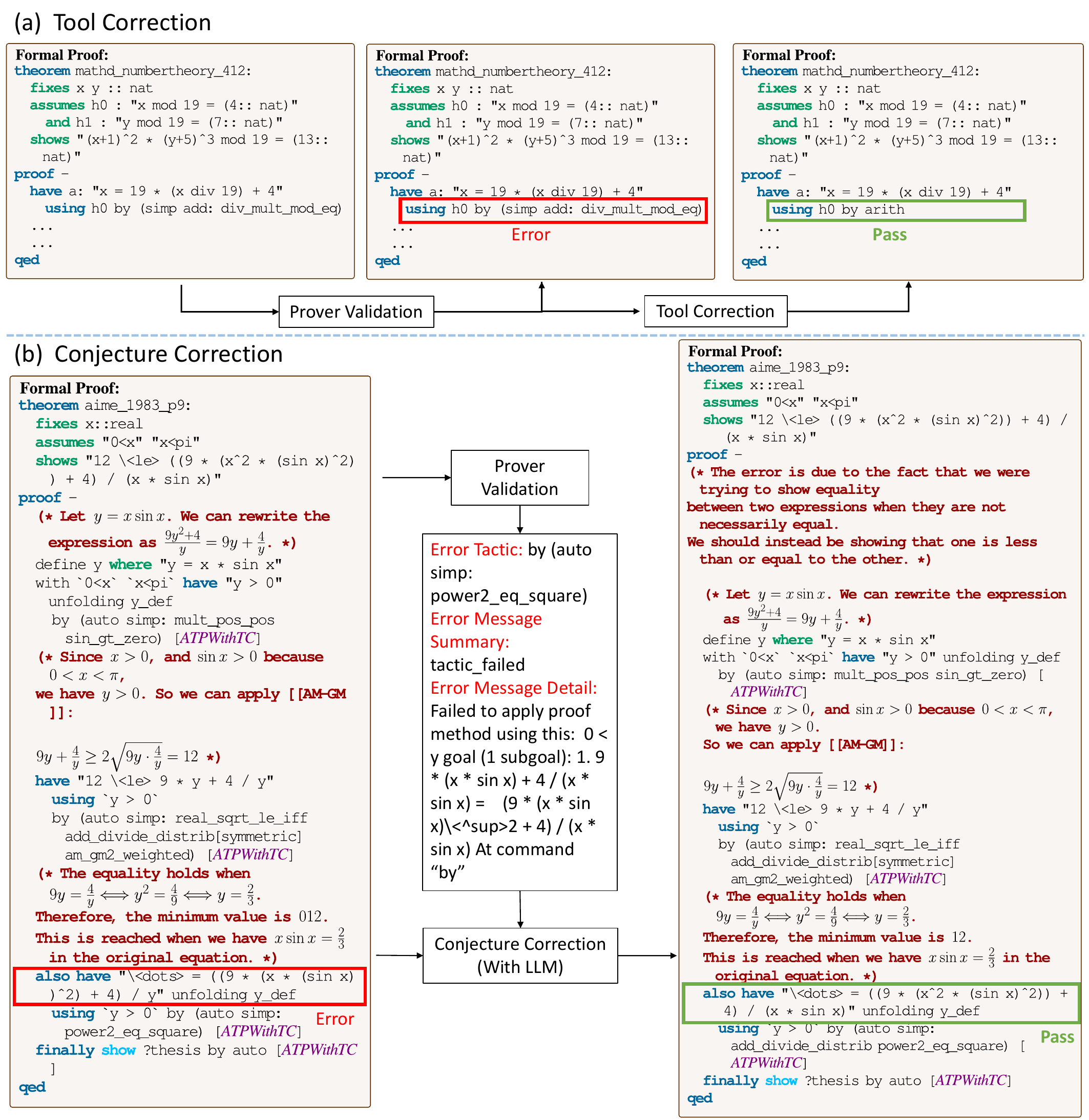}
\caption{
\small
\textbf{Our proposed \shortname framework contains two modules}. \textbf{\methodOneFull:} employ the predefined tools to replace the incorrect tools and prove the conjectures. The prover fails because LLM wrongly believes that $\texttt{by\ (simp\ add:\ div\_mult\_mod\_eq)}$ can prove $x=19*(x~div~19)+4$. Actually, the conjecture is correct and simple, and the prover fails to prove it because it employs an incorrect tool. Hence, the prover successfully proves the conjecture when employing $\texttt{by~arith}$. \textbf{\methodTwoFull:} We design an interaction framework that integrates previous formal sketch and prover error messages to better sketch generation. The steps with the \textcolor{patriarch}{ATPWithTC} delimiters are generated by an automated prover with \methodOneFull.
}
\vspace{-21.5pt}
\label{fig: model figure}

\end{figure}
In this paper, we seek to build \fullname based on LLM, focusing on formal theorem proving. There are two major challenges for LLM generation: 1) hallucination mitigation; 2) interaction with the environment.  To mitigate LLM hallucination, we propose \methodOneFull to leverage prior knowledge and rules to guide incorrect tool replacement. As shown in the observation in Figure~\ref{fig: model figure}, prover fails to prove conjecture $x=19*(x~div~19)+4$ because LLM wrongly believes that $\texttt{by\ (simp\ add:\ div\_mult\_mod\_eq)}$ can prove $x=19*(x~div~19)+4$, while the conjecture is correct but employed tool $\texttt{simp}$ is not powerful enough. \methodOneFull employs predefined tools (e.g. sledgehammer, arith) to guide incorrect tool replacement and finally prove the conjecture. We also propose a general interaction technique with LLM named \methodTwoFull. To further improve and modify the conjectures, \methodTwoFull leverages a general framework that can easily integrate feedback from any environment, in this case, the Isabelle prover, to further polish conjectures. We believe the \shortname presents our insights to mitigate LLM hallucination and interact with the environment.

The proposed method significantly outperforms competing approaches in formal theorem-proving tasks, achieving a pass rate of $51.2\%$ on the miniF2F test dataset, a $5.7\%$ absolute improvement over the previous state-of-the-art. Furthermore, the insights gained from \methodOneFull and \methodTwoFull design can be applied to other frameworks that need to interact with the environment. In summary, our contributions are as follows:
\begin{itemize}[itemsep=5pt,topsep=0pt,parsep=0pt]
  \item We introduce \fullname, a method composed of two components \methodOneFull and \methodTwoFull, to guide automated provers with formal proof sketches.
  \item \methodOneFull employs the predefined
tools to replace the incorrect tools to mitigate hallucination, while \methodTwoFull integrates previous formal sketch and prover error messages to refine proof.
  \item We establish a new SOTA of 55.3\% and 51.2\% on miniF2F validation and test, outperform previous best 7.3\% and 5.7\% respectively. And we newly solve two IMO problems: \texttt{IMO\_1974\_p5} and \texttt{IMO\_1981\_p6}.
\end{itemize}

\section{Related Works}
\textbf{Interactive theorem provers.} 
Contemporary mathematical verification systems are centered on interactive theorem provers (ITPs), including Isabelle \citep{paulson1994isabelle}, Lean \citep{de2015lean}, Coq \citep{barras1997coq}, and Metamath \citep{megill2019computer}. ITPs lay the groundwork for mathematical definitions and theorems on a robust logical foundation through their core kernels. The validation of each theorem is kernel-based and takes place within the ITP. To achieve formal proof, a theorem is initially expressed in the programming language of the ITP and systematically refined into simpler subgoals until it aligns with previously established facts. In this paper, the chosen ATP is Isabelle, known for its potent prover tools, including sledgehammer~\citep{paulson2010three}.

\textbf{Machine learning for formal proving.} Numerous approaches advocate the integration of machine learning with contemporary interactive theorem provers (ITPs) \citep{yang2019learning, gauthier2021tactictoe}. They leverage the recent advancements in language models \citep{polu2020generative, han2021proof, polu2023formal, jiang2022thor, lample2022hypertree, mikula2023magnushammer}. These techniques recommend actions based on the current proving state, and the tree search identifies a sequence of correct steps using actions provided by the language model. Potent methods like MCTS \citep{silver2018general, wu2021tacticzero, laurent2022learning} or dynamic-tree MCTS \citep{wang2023dt} are utilized for this purpose. Previous work \citep{wu2022autoformalization} has demonstrated the few-shot statement autoformalization capability of LLMs \citep{chowdhery2022palm}. In investigating these findings' applicability to proof autoformalization, DSP \citep{jiang2023draft} conducted an in-depth analysis using Draft, Sketch, and Proof. Subgoal-Learning \citep{zhao2023decomposing} further employs a subgoal-goal-based informal proof approach. In an effort to support the open-source community, LeanDojo \citep{yang2023leandojo} created a Lean playground that includes toolkits, data, models, and benchmarks. 
While these methods directly use the results generated by LLMs, we adopt a different approach by employing predefined tools to post-process the generations to mitigate hallucination, specifically \methodOneFull.

\textbf{Large language model refinement.}
Calibration studies conducted on LLLMs reveal that the probabilistic predictions made by current LLMs are closely aligned with the actual frequencies of token occurrences, resulting in well-calibrated predictions for specific tasks~\citep{guo2017calibration,kadavath2022language,jiang2020can}. As LLMs exhibit reliable calibration, an increasing number of research studies emphasize using self-evaluation for verification. For instance, Reflexion~\citep{shinn2023reflexion} leverages an agent with dynamic memory and self-reflection capabilities, while Self-Refine~\citep{madaan2023selfrefine} proposes a method to generate outputs from LLMs and refine their previously generated outputs based on their own feedback. Taking a similar approach, methods like Self-Debug~\citep{chen2023teaching} and CRITICS~\citep{gou2023critic} interact with code interpreters to further debug. In contrast, Progressive-Hint Prompting~\citep{zheng2023progressive} iteratively extracts hints from previous LLM's answers as hints for the next answer generation. However, previous works require extensive prompts, including generation prompts and refine prompts. Our approach \methodTwoFull  refines generation with instruction but does not
collect paired (generation, error \& refinement) prompts.

\section{Method}
This section describes our \fullname for formal proof automation, which leverages \methodOneFull and \methodTwoFull to guide automated formal theorem provers with proof sketches. 

\subsection{Background: Pipeline of DSP}
 DSP~\citep{jiang2023draft} aims to generate a formal sketch from an informal statement, verifiable by an off-the-shelf automated theorem prover. DSP creates $N$ demonstration examples, denoted as $E = {E_1, E_2, ..., E_N }$, each containing informal/formal components (statements, proofs, sketches). The pipeline of DSP has the following three steps.

\textbf{Informal proof generation.} There are two scenarios: one with an existing human informal proof and another where a language model generates draft-proof candidates without a human reference. For LLM-generated informal proof, DSP provides the model with a few examples containing both (statement, informal proof) for informal proof generation. Subsequently, DSP presents a problem statement that needs to be translated and the model then generates the subsequent tokens to produce the desired informal proof.

\textbf{Formal proof generation.} DSP leverages the few-shot learning capabilities of a large language model. Specifically, DSP provides the model with a few example pairs containing (statement, informal proof, formal sketch) for formal proof generation. Subsequently, DSP presents a (statement, informal proof) that needs to be translated. The model then generates the subsequent tokens to produce the desired formal sketch.

\textbf{Prover validation.} In the final phase, off-the-shelf automated provers address sketch gaps. These systems create formally valid proofs. DSP framework remains agnostic to prover type (symbolic, neural, hybrid). Successful prover results yield verifiable formal proofs.
\subsection{Tool Correction}
\begin{algorithm}[t]
\caption{Pseudocode of \methodOneFull in a Python-like style.}
\label{alg:toolCorrection}
\small

    
        

\definecolor{codeblue}{rgb}{0.25,0.5,0.5}
\definecolor{codekw}{rgb}{0.85, 0.18, 0.50}
\lstset{
  backgroundcolor=\color{white},
  basicstyle=\fontsize{7.5pt}{7.5pt}\ttfamily\selectfont,
  columns=fullflexible,
  breaklines=true,
  captionpos=b,
  commentstyle=\fontsize{7.5pt}{7.5pt}\color{codeblue},
  keywordstyle=\fontsize{7.5pt}{7.5pt}\color{codekw},
  escapechar={|}, 
  showstringspaces=false,
}
\begin{lstlisting}[language=python]
#tactic_list: list of the tactics of formal proof
#prover: Isabelle Prover 
#TCUsage: whether employ Tool Correction
tool_heuristics=['by auto','by arith','by blast', 'by simp',
'by fastforce', 'by force', 'by eval', 'by presburger', 'by sos',
'by linarith', 'by (auto simp: field_simps)', 'sledgehammer']
for tactic in tactic_list:
    use_heuristics=False
    output = prover.run_tac(tactic)
    if not (output['error'] is None):
        if TCUsage: # Use Tool Correction or Not
            if tactic.strip().startswith("by") or tactic.strip()==("."):
                use_heuristic=True
    
        if ("sledgehammer" in tactic) or use_heuristic:
            for tool_try in tool_heuristics:
                output = prover.run_tac(tool_try)
                if output['error'] is None:
                    break
    if output['error'] is not None:
        return "tactic_failed", output
    if output['tactic_state'] == 'no goals':
        return "success", output
        
return "proof_incomplete", output
\end{lstlisting}

\end{algorithm}

\methodOneFull employs prior knowledge to employ predefined tools (e.g. sledgehammer) to guide incorrect tool replacement, as shown in Algorithm \ref{alg:toolCorrection}. We introduce the \methodOneFull as a remedy to alleviate the generation errors stemming from Large Language Models (LLMs). Through empirical observation, it becomes evident that despite the factual accuracy of conjectures, LLMs at times adopt misguided tools that do not withstand validation by theorem provers, as shown in Figure \ref{fig: model figure}.

For instance, consider the statement $x=19*(x~div~19)+4$, where LLM proposes to utilize the tactic $\texttt{by\ (simp\ add:\ div\_mult\_mod\_eq)}$, leading to failure. This is the LLM hallucination, as $\texttt{by\ (simp\ add:\ div\_mult\_mod\_eq)}$ is suited for proving $a=a~div~b~*~b + a~mod~b$ but not $x=19*(x~div~19)+4$. Substituting it with $\texttt{by~arith}$ enables the theorem prover to successfully verify $x=19*(x~div~19)+4$. Hence, in certain instances, LLM might formulate correct conjectures but employ inappropriate tools, resulting in unsuccessful proof attempts. To address this, \methodOneFull leverages predefined tools to enhance the success rate.

The \methodOneFull approach entails the validation of a given tactic $t$ using Isabelle. If validation succeeds, we proceed; if not, \methodOneFull intervenes to modify the tactic. Specifically, when a tactic is equal to ``." or commencing with ``by" or ``sledgehammer" but the tactic fails, we attempt the application of $t_{tool}$. This $t_{tool}$ can be either: 1) ``sledgehammer" or; 2) by $\texttt{tool}$ with $\texttt{tool}$ belonging to the set {(\texttt{auto}, \texttt{simp}, \texttt{blast}, \texttt{fastforce}, \texttt{force}, \texttt{eval}, \texttt{presburger}, \texttt{sos}, \texttt{arith}, \texttt{linarith}, \texttt{auto simp: field simps})}.

By integrating \methodOneFull, we systematically explore the applicability of ``sledgehammer" and 11 heuristic tools. If any of these successfully pass the theorem prover, we progress to the subsequent tactics. However, if proof still fails to prove the tactic after trying all $t_{tool}$ fail, the overall proof attempt is deemed unsuccessful.
\begin{algorithm}[t]
\caption{Pseudocode of \methodTwoFull in a Python-like style.}
\label{alg:chat}
\small

\definecolor{codeblue}{rgb}{0.25,0.5,0.5}
\definecolor{codekw}{rgb}{0.85, 0.18, 0.50}
\lstset{
  backgroundcolor=\color{white},
  basicstyle=\fontsize{7.5pt}{7.5pt}\ttfamily\selectfont,
  columns=fullflexible,
  breaklines=true,
  captionpos=b,
  commentstyle=\fontsize{7.5pt}{7.5pt}\color{codeblue},
  keywordstyle=\fontsize{7.5pt}{7.5pt}\color{codekw},
  escapechar={|}, 
  showstringspaces=false,
}
\begin{lstlisting}[language=python]
#round_count: the current round number
#prompt_sample: the prompt and propsoed question
#previous_response: previous formal proof
#error_info:error informalion from Isabelle
input=[{"role": "system",  "content": "You are an expert in \
Mathematical Proof and Isabelle Proof Assistant. Follow the given \ 
examples and complete the proof with Isabelle Proof Assistant"},
{"role": "user", "content": prompt_sample}]
if round_count%5!=0: #If Falase, the initial round.
#Otherwise, then Refine round.
 #Refine Rround
  input.append({"role": "assistant", "content": previous_respon})
  input.append({"role": "user", "content": "(*The last proof has the \
  following errors from Isabelle Prover. Therefore,\n 1) Please Follow  \
  the Above Prompt;\n\n 2) And Utilize the Following Errors to redo \
  the last formal proof.\n {}.\n\n*)\n\n \
  proof -\n".format(error_info)})
json_obj = openai.ChatCompletion.create(messages=input)
\end{lstlisting}
\end{algorithm}
\subsection{Conjecture Correction}
For \methodTwoFull, we design a framework that can easily integrate previous formal sketches and error messages from the prover to improve sketch generation. LLMs, particularly GPT-4, can leverage prior responses or contextual cues for improved output. Nonetheless, integrating feedback into mathematical proving remains a challenge. This stems from two primary factors: 1) diverse theorem provers employ distinct syntax, complicating the design of varied prompts; 2) often require an extensive token count, incurring a high computational cost and exceeding model length limits. To address these limitations, \shortname uses \methodTwoFull, offering a versatile refinement pipeline that can transform a non-refined framework into a refined one.  Compared to the previous refinement framework, such as Self-Refine~\citep{madaan2023selfrefine} or Self-Debug~\citep{chen2023teaching}, the proposed \methodTwoFull refines generation with instruction, but does not collect paired (generation, error \& refinement) prompts. The details are shown in Algorithm~\ref{alg:chat}.

\textbf{Initial round generation.}  In the initial round generation, we follow the same process as DSP, directly producing informal or formal proofs without prover error messages.

\textbf{Rectification round.} Our approach also involves the use of an LLM (e.g. GPT-4 \citep{bubeck2023sparks}) for rectification. In contrast to the initial round generation, the rectification employs the same initial prompt as the first round but appends error messages from the prover. As all formal proof begins with \texttt{proof -}, we add \texttt{proof -} at the end of the instruction so that the LLM response is formal proof.

\textbf{Reset initial round generation.} Since the rectification round builds upon the first round, the quality of \methodTwoFull performance is tied to the initial round proof. To ensure that a potentially subpar initial round proof does not negatively affect subsequent proofs, we regenerate the initial round proof at interaction rounds $K$, $2K$, $3K$ and so on, refining its generation in the remaining rounds.  For example, when working with 200 attempts and setting $K$ to 5, \methodTwoFull partitions the 200 attempts into 40 patches. Each patch consists of the first proof derived from DSP, followed by four subsequent refined proofs that build upon the previous proof and incorporate the error message provided by the prover.

\section{Experiment}
\subsection{Dataset}
In this study, we assess our approach using the miniF2F dataset~\citep{zheng2021minif2f}, which is a collection of $488$ formal mathematical problems derived from high-school competitions and expressed in three formal languages: Lean~\citep{de2015lean}, HOL-Light~\citep{bansal2019holist}, and Isabelle~\citep{paulson1994isabelle}. The dataset is divided into validation and test sets, each containing $244$ problems. These problems are sourced from three distinct categories, with $260$ problems extracted from the MATH dataset~\citep{hendrycks2021measuring}, $160$ problems taken from actual high school mathematical competitions (AMC, AIME, and IMO), and $68$ problems specially crafted to mirror the difficulty level of the aforementioned competitions.

\textbf{Evaluation.} The objective of our study is to generate formal sketches for the problems in the miniF2F dataset. We consider a proof valid if and only if (a) it does not have any "cheating" keywords (\texttt{sorry} and \texttt{oops}) that terminate a proof without completion, and (b) Isabelle must be capable of verifying the corresponding formal statement with the proof.

\textbf{Implementation details.} In our research, we utilized GPT-4 as the Language Model Model (LLM) for generating informal drafts and formal sketches. The temperature of GPT-4 was set to 0.7, with 200 attempts. The details of baselines are shown in Appendix.

\subsection{Main Results}
\begin{table*}[t]
\begin{center}
\caption{
\small
\textbf{Proving success rates on the \miniff dataset with Isabelle.} 
The table displays the success rates of previous works and the \shortname, using both human and GPT-4 informal proofs. The highest success rates for each set are highlighted in bold.
}
\label{tab:main_results} 
\small
\begin{tabular}{lcc}
    \toprule
    Success rate & \miniff-valid & \miniff-test\\
    \midrule
    \multicolumn{1}{l}{\textit{Baselines}} \\
    \midrule
    Sledgehammer~\citep{paulson2010three}              & $9.9\%$ & $10.4\%$ \\
    Sledgehammer + heuristics~\citep{jiang2023draft} & $18.0\%$ & $20.9\%$ \\
    Thor~\citep{jiang2022thor} & $28.3\%$ & $29.9\%$ \\
    Thor + expert iteration~\citep{wu2022autoformalization} & $37.3\%$ & $35.2\%$ \\
    \midrule
    \multicolumn{1}{l}{\textit{Draft, Sketch, and Prove (100 attempts)}}~\citep{jiang2023draft} \\
    \midrule
    Human informal proof    & $42.6\%$ & $39.3\%$ \\
    $540$B Minerva informal proof & $42.6\%$ & $38.9\%$ \\
    \midrule
     \textit{Subgoal-Learning (100 attempts)}~\citep{zhao2023decomposing}   & $48.0\%$ & $45.5\%$ \\
    \midrule
    \multicolumn{1}{l}{\textit{\shortname (Ours)}} \\
    \midrule
    GPT-4 informal proof ~(100 attempts)  & $52.8\%$ & $44.2\%$ \\
    GPT-4 informal proof ~(200 attempts)  & $54.9\%$ & $47.9\%$ \\
    Human informal proof (100 attempts)   & $52.0\%$ & $47.1\%$ \\
    Human informal proof (200 attempts)   & $\mathbf{55.3}\%$ & $\mathbf{51.2\%}$ \\
   
    \bottomrule
\end{tabular}
\end{center}
\vspace{-0.1in}
\end{table*}
Table \ref{tab:main_results} presents the distribution of successful formal proofs obtained from the miniF2F dataset using the interactive theorem prover Isabelle. An examination of the results presented in Table \ref{tab:main_results} reveals a conspicuous enhancement in the efficacy of the Sledgehammer automated prover, owing to the integration of $11$ supplementary heuristic tactics~\citep{jiang2023draft}. Noteworthy achievements are also realized through deploying the DSP-based methods (DSP and Subgoal), attaining success rates of $39.3\%$ and $45.5\%$, respectively on the miniF2F test set.

By harnessing informal proofs generated by GPT-4, our proposed method achieves success rates of $54.9\%$ and $47.9\%$ on the validation and test sets of miniF2F respectively. This performance persists even when the attempt number is set at $100$, affirming its robustness. When the attempt number is $100$, compared to 540B Minerva informal proof with DSP, our proposed \shortname improves the performance on miniF2F validation set from $42.6\%$ to $52.8\%$ and miniF2F test set from $38.9\%$ to $44.2\%$. This outcome can be attributed to the \methodOneFull and \methodTwoFull.

In instances where human informal proofs are employed, our proposed method demonstrates impressive success rates of $55.3\%$ and $51.2\%$ on the validation and test sets of miniF2F. Comparative analysis against DSP reveals an improvement of $12.7\%$ and $11.9\%$ on the validation and test sets respectively for miniF2F. Furthermore, when contrasted with the previous state-of-the-art Subgoal-Learning model, our approach showcases an advancement of $7.3\%$ and $5.7\%$ on the miniF2F validation and test sets respectively. 

The performance of human informal proofs surpasses that of GPT-4 generated counterparts, especially on the test set. This substantiates the notion that precision in informal proofs is important for generating formal sketches.
\subsection{Ablation Study}
\begin{table*}[t]
\begin{center}
\caption{
\small
\textbf{Ablation results on the miniF2F dataset with Isabelle.} There are three important conclusions: 1) GPT-4 is better than Codex for mathematical proving; 2) \methodOneFull can consistently improve performance; 3) \methodTwoFull can improve performance but needs more attempts. \textbf{Our proposed method degrades to DSP~\citep{jiang2023draft} when without \methodOneFull and \methodTwoFull}.
}
\label{tab:ablation_study} 
\small
\begin{tabular}{ccccccc}
    \toprule
    Attempt &  Formal Proof &  Informal Proof& \methodOne & \methodTwo & miniF2F-valid & miniF2F-test\\

    \midrule
     \multirow{6}{*}{100}   &Codex  &$540$B Minerva&$\xmark$ & $\xmark$ & $42.6\%$ & $38.9\%$ \\
     &  GPT-4 &   GPT-4 & $\xmark$ & $\xmark$ & $48.3\%$ & $38.9\%$ \\
     \cmidrule{2-7}
      &  Codex &  Human & $\xmark$ & $\xmark$ & $42.6\%$ & $39.3\%$ \\
       &  GPT-4 &    Human & $\xmark$ & $\xmark$ & $47.9\%$ & $39.7\%$ \\
     \cmidrule{2-7}

     &  GPT-4 &   GPT-4 &$\cmark$ & $\cmark$& $52.8\%$ & $44.2\%$ \\
     &   GPT-4 &   Human &$\cmark$ & $\cmark$ & $52.0\%$ & $47.1\%$ \\
    \midrule

  \multirow{8}{*}{200} & GPT-4 &   GPT-4 & $\xmark$ & $\xmark$ & $49.5\%$ & $40.9\%$ \\
   & GPT-4 &    GPT-4 & $\cmark$ & $\xmark$ & $55.3\%$ & $45.0\%$ \\
    &  GPT-4 &    GPT-4 & $\xmark$ & $\cmark$ & $48.3\%$ & $40.9\%$ \\
     &  GPT-4 &    GPT-4 & $\cmark$ & $\cmark$ & $54.9\%$ & $47.9\%$ \\
     \cmidrule{2-7}

      &    GPT-4 &    Human & $\xmark$ & $\xmark$ & $50.4\%$ & $42.6\%$ \\
  &   GPT-4 &    Human & $\cmark$ & $\xmark$ & $52.8\%$ & $45.9\%$ \\
   &  GPT-4 &    Human & $\xmark$ & $\cmark$ & $46.7\%$ & $43.0\%$ \\
    &   GPT-4 &    Human & $\cmark$ & $\cmark$ & $\mathbf{55.3\%}$ & $\mathbf{51.2\%}$ \\

    \bottomrule
\end{tabular}
\end{center}
\vspace{-0.1in}
\end{table*}
\begin{figure}[t]
\setlength{\abovecaptionskip}{0.1cm}
\centering
\includegraphics[width=0.45\textwidth]{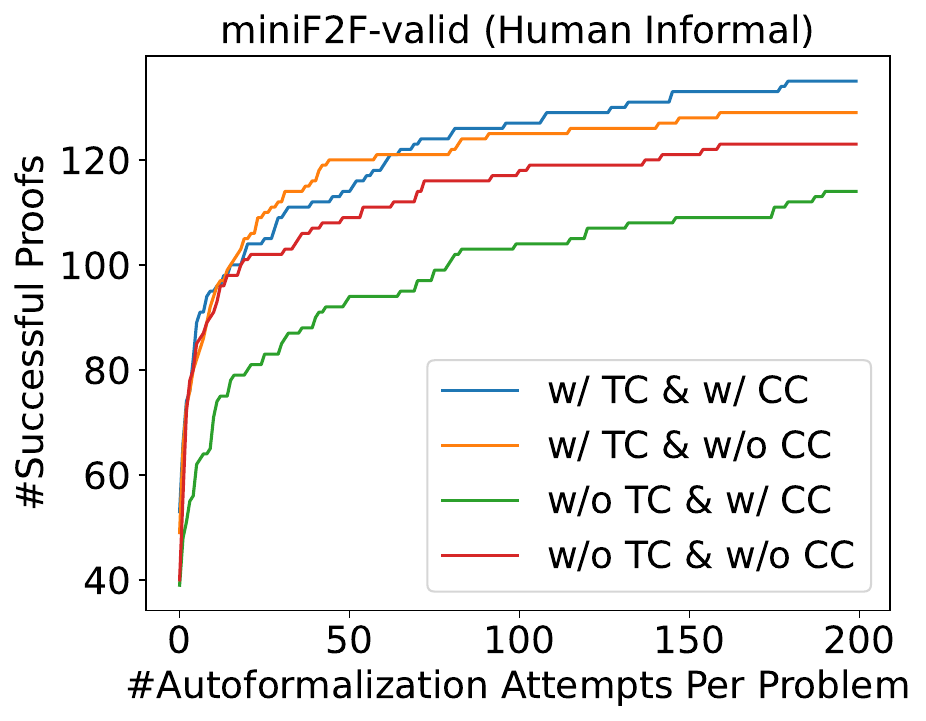}
\hspace{0in}
\includegraphics[width=0.45\textwidth]{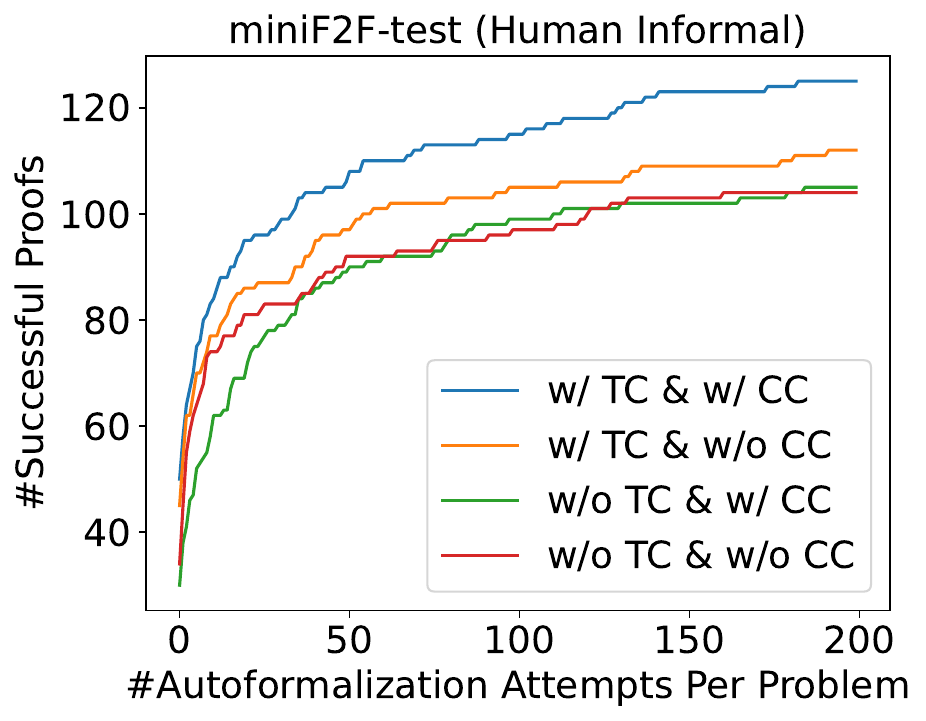}
\hspace{0in}
\includegraphics[width=0.45\textwidth]{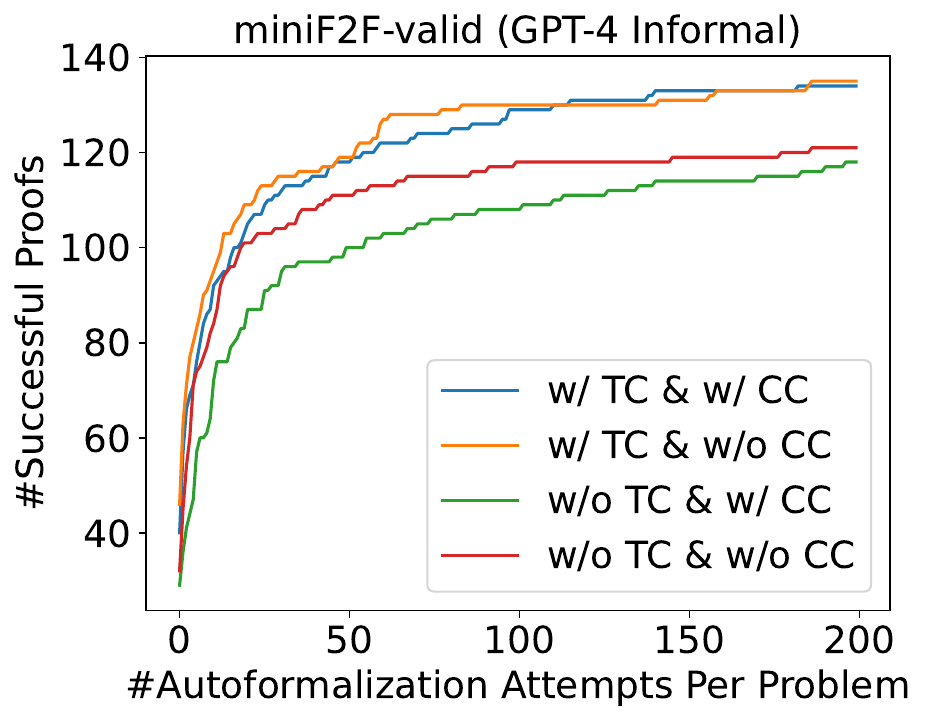}
\hspace{0in}
\includegraphics[width=0.45\textwidth]{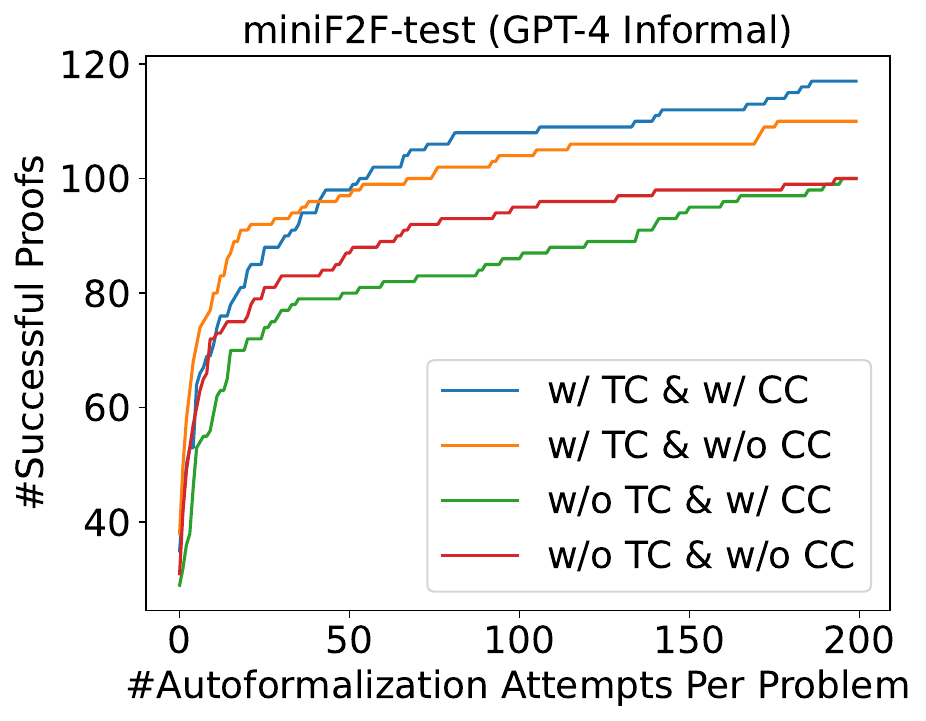}
\caption{
\small
\textbf{Number of problems solved on miniF2F against the number of autoformalization attempts per problem}. On miniF2F validation and test set, we have shown the results of \methodOneFull (\methodOne) and \methodTwoFull (\methodTwo) on human informal proof and GPT-4 informal proof respectively.
}
\label{fig: ablation}
\vspace{-15pt}
\end{figure}
\textbf{GPT-4 is better than Codex, especially on miniF2F validation dataset..} In the absence of \methodOneFull and \methodTwoFull, our proposed method experiences degradation to DSP. Referring to  Table~\ref{tab:ablation_study}, when considering the informal proof generated by LLM (GPT-4 or 540B Minerva), GPT-4 is better than Codex~\citep{chen2021evaluating}. When compared with the deployment of Codex for generating formal sketches, GPT-4 demonstrates improvements of $5.3\%$ and $0.4\%$ on the validation and test subsets of miniF2F, respectively, while utilizing the same attempt number $100$ and human informal proof. This substantiates the notion that GPT-4 indeed enhances performance. 

\textbf{\methodOneFull: consistently improve performance.} As evident from Table~\ref{tab:ablation_study} and Figure~\ref{fig: ablation}, the inclusion of \methodOneFull yields enhanced performance. Similarly,  when assessing GPT-4-generated informal proofs on the miniF2F test set, \methodOneFull elicits improvements of $4.1\%$ and $7.0\%$ in the absence and presence of \methodTwoFull, respectively. When considering human informal proofs on the miniF2F test set, \methodOneFull showcases enhancements of $3.3\%$ and $8.2\%$ in scenarios devoid of and accompanied by \methodTwoFull, respectively. Therefore, regardless of whether the informal sketch is generated by GPT-4 or created manually by a human, \methodOneFull consistently enhances performance and can further benefit from the addition of \methodTwoFull.

\textbf{\methodTwoFull: further improves performance, prefers more powerful prover and requires more attempts to be convergent.} The outcomes presented in Table~\ref{tab:ablation_study} and illustrated in Figure~\ref{fig: ablation} underscore the efficacy of integrating \methodTwoFull, albeit at the expense of requiring an increased number of attempts to achieve convergence. When considering
human informal proofs on the miniF2F test set, \methodTwoFull showcases enhancements of 0.4\%
and 5.3\% in scenarios devoid of and accompanied by \methodOneFull, respectively. This suggests that \methodTwoFull improves proof quality, but needs a more powerful prover (e.g. with \methodOneFull) to fill the formal gaps. \methodTwoFull needs more attempts to be convergent because \methodTwoFull modifies the initial proof to generate subsequent proofs, which strongly hinges on the quality of the initial proof. Specifically, \methodTwoFull partitions the pool of 200 attempts into 40 patches, wherein the first proof originates from DSP, and the subsequent four are based on the initial proof. Furthermore, it's worth noting that, in theory, any problems solvable through DSP remain solvable using our approach, as DSP is equivalent to our initial proof generation without \methodOneFull.

\textbf{Attempt number: \shortname benefits more with attempt number increment.} In the absence of \methodOneFull and \methodTwoFull, our proposed method reduces to DSP. Within the validation set with human informal proofs, when the number of attempts is escalated from $100$ to $200$ (shown in Table \ref{tab:ablation_study}), the performance of DSP experiences a gain from $47.9\%$ to $50.4\%$, achieving a $2.5\%$ improvement. Conversely, our proposed approach exhibits a performance improvement from $52.0\%$ to $55.3\%$, reflecting a more substantial $3.3\%$ enhancement. For the test set, DSP's performance improves from $39.7\%$ to $42.6\%$, marking a $2.9\%$ increment. In contrast, our method demonstrates an increment from $47.1\%$ to $51.2\%$, indicating a more $4.1\%$ boost. This divergence implies that our proposed approach effectively surpasses the performance limitations of DSP, highlighting the potential efficacy of expanding the attempt number to further enhance performance differences.

\begin{figure}[t]
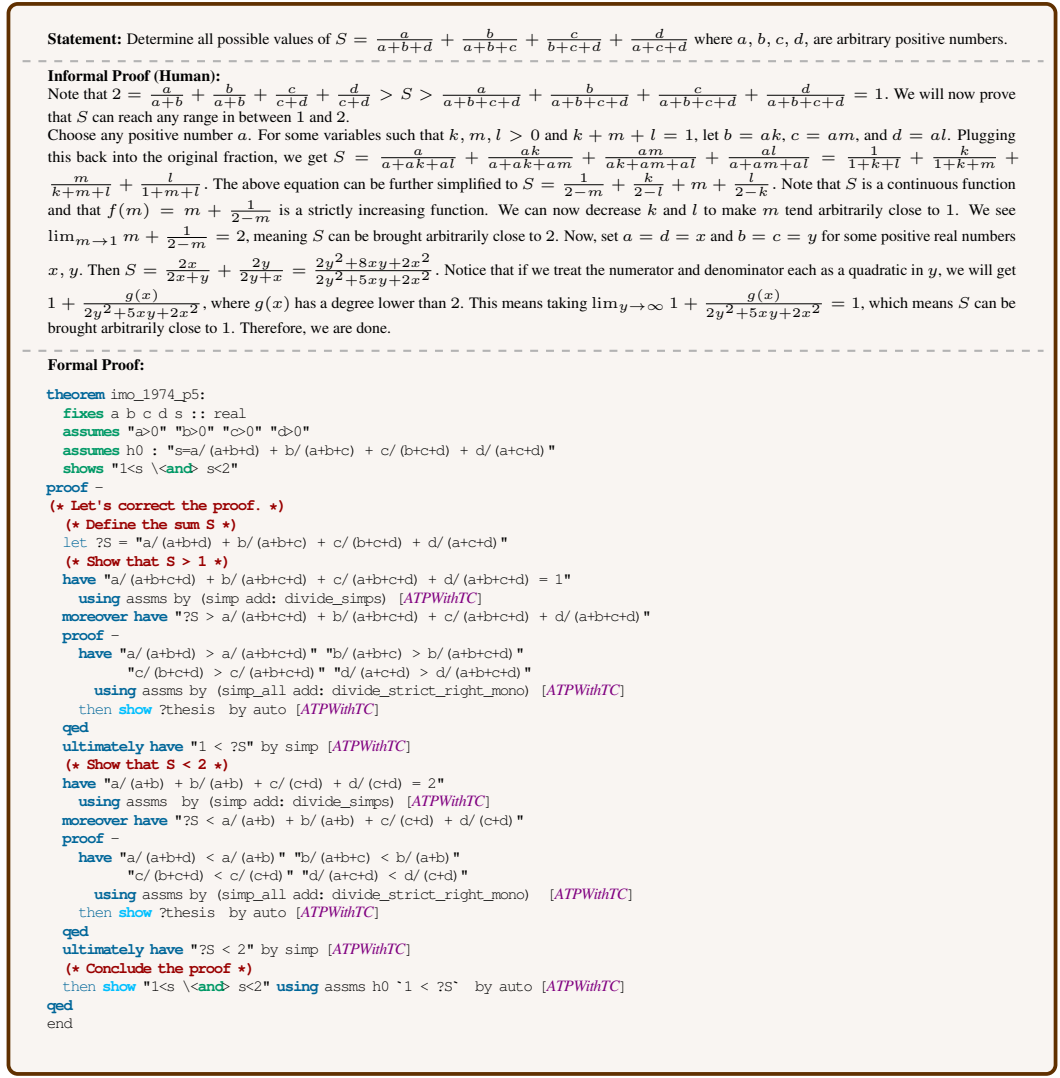

\setlength{\abovecaptionskip}{0.1cm}
\begin{tcolorbox}[colback=mybrown!5!white,colframe=mybrown!75!black]
\begin{tiny}
\textbf{Statement:}
Determine all possible values of $S = \frac{a}{a+b+d}+\frac{b}{a+b+c}+\frac{c}{b+c+d}+\frac{d}{a+c+d}$ where $a, b, c, d,$ are arbitrary positive numbers.
\Sepline
\textbf{Informal Proof (Human):}\\
Note that $2 = \frac{a}{a+b}+\frac{b}{a+b}+\frac{c}{c+d}+\frac{d}{c+d} > S > \frac{a}{a+b+c+d}+\frac{b}{a+b+c+d}+\frac{c}{a+b+c+d}+\frac{d}{a+b+c+d} = 1.$ We will now prove that $S$ can reach any range in between $1$ and $2$.

Choose any positive number $a$. For some variables such that $k, m, l > 0$ and $k + m + l = 1$, let $b = ak$, $c = am$, and $d = al$. Plugging this back into the original fraction, we get 
$S = \frac{a}{a+ak+al}+\frac{ak}{a+ak+am}+\frac{am}{ak+am+al}+\frac{al}{a+am+al} = \frac{1}{1+k+l}+\frac{k}{1+k+m}+\frac{m}{k+m+l}+\frac{l}{1+m+l}.$
The above equation can be further simplified to 
$S = \frac{1}{2-m}+\frac{k}{2-l}+m+\frac{l}{2-k}.$
Note that $S$ is a continuous function and that $f(m) = m + \frac{1}{2-m}$ is a strictly increasing function. We can now decrease $k$ and $l$ to make $m$ tend arbitrarily close to $1$. We see $\lim_{m\to1} m + \frac{1}{2-m} = 2$, meaning $S$ can be brought arbitrarily close to $2$. 
Now, set $a = d = x$ and $b = c = y$ for some positive real numbers $x, y$. Then 
$S = \frac{2x}{2x+y} + \frac{2y}{2y+x} = \frac{2y^2 + 8xy + 2x^2}{2y^2 + 5xy + 2x^2}.$
Notice that if we treat the numerator and denominator each as a quadratic in $y$, we will get $1 + \frac{g(x)}{2y^2 + 5xy + 2x^2}$, where $g(x)$ has a degree lower than $2$. This means taking $\lim_{y\to\infty} 1 + \frac{g(x)}{2y^2 + 5xy + 2x^2} = 1$, which means $S$ can be brought arbitrarily close to $1$. Therefore, we are done.
$ $
\Sepline
\textbf{Formal Proof:}
\vspace{-1.5em}

\begin{lstlisting}[style=isabelle]
theorem imo_1974_p5:
  fixes a b c d s :: real
  assumes "a>0" "b>0" "c>0" "d>0"
  assumes h0 : "s=a/(a+b+d) + b/(a+b+c) + c/(b+c+d) + d/(a+c+d)"
  shows "1<s \<and> s<2"
proof -
(* Let's correct the proof. *)
  (* Define the sum S *)
  let ?S = "a/(a+b+d) + b/(a+b+c) + c/(b+c+d) + d/(a+c+d)"
  (* Show that S > 1 *)
  have "a/(a+b+c+d) + b/(a+b+c+d) + c/(a+b+c+d) + d/(a+b+c+d) = 1" 
    using assms by (simp add: divide_simps) [ATPWithTC]
  moreover have "?S > a/(a+b+c+d) + b/(a+b+c+d) + c/(a+b+c+d) + d/(a+b+c+d)" 
  proof -
    have "a/(a+b+d) > a/(a+b+c+d)" "b/(a+b+c) > b/(a+b+c+d)" 
          "c/(b+c+d) > c/(a+b+c+d)" "d/(a+c+d) > d/(a+b+c+d)"
      using assms by (simp_all add: divide_strict_right_mono) [ATPWithTC]
    then show ?thesis  by auto [ATPWithTC]
  qed
  ultimately have "1 < ?S" by simp [ATPWithTC]
  (* Show that S < 2 *)
  have "a/(a+b) + b/(a+b) + c/(c+d) + d/(c+d) = 2" 
    using assms  by (simp add: divide_simps) [ATPWithTC]
  moreover have "?S < a/(a+b) + b/(a+b) + c/(c+d) + d/(c+d)"
  proof -
    have "a/(a+b+d) < a/(a+b)" "b/(a+b+c) < b/(a+b)" 
          "c/(b+c+d) < c/(c+d)" "d/(a+c+d) < d/(c+d)"
      using assms by (simp_all add: divide_strict_right_mono)  [ATPWithTC]
    then show ?thesis  by auto [ATPWithTC]
  qed
  ultimately have "?S < 2" by simp [ATPWithTC]
  (* Conclude the proof *)
  then show "1<s \<and> s<2" using assms h0 `1 < ?S`  by auto [ATPWithTC]
qed
end

\end{lstlisting}
\end{tiny}
\end{tcolorbox}
\caption{
\small
\textbf{A successful formal proof synthesized with human informal proof}. With \methodOneFull and \methodTwoFull, we successfully solve an IMO problem \texttt{IMO\_1974\_p5}. The steps with the \textcolor{patriarch}{ATPWithTC} delimiters are generated by an automated prover with \methodOneFull. We also solve \texttt{IMO\_1959\_p1} with GPT-4 informal proof, which is shown in the Appendix. 
}
\vspace{-20pt}
\label{fig:case_study}
\end{figure}

\subsection{Case Study}

We solve another IMO problem \texttt{IMO\_1959\_p1} with GPT-4 informal proof, which is also solved via DSP with 540B Minerva. Furthermore, to present the effectiveness of our method, we provide a formal sketch of an IMO problem named \texttt{IMO\_1974\_p5} that remains unproven by earlier state-of-the-art methods. As demonstrated in Figure~\ref{fig:case_study}, our \shortname successfully proves \texttt{IMO\_1974\_p5} with \methodOneFull and \methodTwoFull. We have shown the interaction details of \texttt{IMO\_1974\_p5} and \texttt{IMO\_1959\_p1} in the Appendix. 

\section{Conclusion}
In this paper, we introduced \fullname, a novel pipeline that takes advantage of \methodOneFull and \methodTwoFull. \methodOneFull employs prior knowledge to employ predefined tools (e.g. sledgehammer) to guide incorrect tool replacement. \methodTwoFull, interacting with the prover environment, integrates previous formal sketch and prover error messages for better sketch generation.  We demonstrated the feasibility and effectiveness of \shortname by reaching state-of-the-art performance 55.3\% and 51.2\% on the miniF2F dataset validation and test, respectively, with the Isabelle theorem prover. Central to our method is the incorporation of prior knowledge and the development of a comprehensive GPT-4 refinement framework. Our ablations showed that both \methodOneFull and \methodTwoFull are critical to the success of \shortname.

\nocite{*}
\bibliography{iclr2024_conference}
\bibliographystyle{iclr2024_conference}

\clearpage

\newpage
\appendix
\section*{\Large{Appendix}}

\lstdefinelanguage{isabelle}{%
    keywords=[1]{type_synonym,datatype,fun,abbreviation,definition,proof,lemma,theorem,qed,corollary,have,hence,also,finally,ultimately,moreover,using,\{},
    keywordstyle=[1]\bfseries\color{isarblue},
    keywords=[2]{where,assumes,shows,fixes,and},
    keywordstyle=[2]\bfseries\color{isargreen},
    keywords=[3]{if,then,else,case,SOME,let,in,O},
    keywordstyle=[3]\color{isarblue},
    keywords=[4]{ATPWithTC},
    keywordstyle=[4]\it\color{patriarch},
    keywords=[5]{show,assume,obtain},
    keywordstyle=[5]\bfseries\color{isarfaintblue},
}

\lstdefinestyle{isabelle}{%
  language=isabelle,
  escapeinside={\&}{&},
  columns=fixed,
  extendedchars,
  basewidth={0.5em,0.45em},
  basicstyle=\singlespacing\ttfamily\small,
  mathescape,
  morecomment=[s][\bfseries\color{red}]{(*}{*)},
  morecomment=[l][\bfseries]{####},
  breaklines=true,
}


\section{Baseline}
To evaluate the effectiveness of our proposed methodology, we employed several baseline methods, which follow previous work setting~\citep{jiang2023draft,zhao2023decomposing}.

\textbf{Sledgehammer with heuristics} The first baseline is Sledgehammer~\citep{paulson2010three}, which a proof automation tool in the Isabelle environment. Additionally, we utilized Sledgehammer supplemented with heuristics, which integrates 11 prevalent tactics (i.e., auto, simp, blast, fastforce, force, evapl, presburger, sos, arith, linarith, auto simp: field simps) with Sledgehammer. If all tactics fail, the system employs Sledgehammer~\citep{jiang2023draft}.

\textbf{Language model based methods}  Thor~\citep{jiang2022thor} combines language models with automatic theorem provers to hep select premises from a vast library. Thor+expert~\citep{wu2022autoformalization} iteration enhances a neural theorem prover by training it on theorems that have been successfully formalized. Draft, Sketch, and Prove (DSP)~\citep{jiang2023draft} transforms informal proofs into formal sketches and utilizes these formal sketches to guide an automated prover. Another LLM-based method is Subgoal-Proof Learning~\citep{zhao2023decomposing}, which incorporates subgoal proof to replace informal proof and proposes a prompt selection framework.

Following previous work ~\citep{jiang2023draft,zhao2023decomposing}, we excluded representative methods such as HyperTree Proof Search (HTPS)~\citep{lample2022hypertree} and GPT-f with expert iteration~\citep{polu2023formal}, which are implemented using Lean~\cite{de2015lean}, a different interactive theorem prover. The disparity in tactics and automation between Lean and Isabelle renders them not directly comparable to our method.

\clearpage

\section{IMO 1974 p5 Case}
\begin{figure}[H]
\begin{tcolorbox}[colback=mybrown!5!white,colframe=mybrown!75!black]
\begin{small}
\textbf{Statement:}
Determine all possible values of $S = \frac{a}{a+b+d}+\frac{b}{a+b+c}+\frac{c}{b+c+d}+\frac{d}{a+c+d}$ where $a, b, c, d,$ are arbitrary positive numbers.
\Sepline
\textbf{Informal Proof (Human):}\\
Note that $2 = \frac{a}{a+b}+\frac{b}{a+b}+\frac{c}{c+d}+\frac{d}{c+d} > S > \frac{a}{a+b+c+d}+\frac{b}{a+b+c+d}+\frac{c}{a+b+c+d}+\frac{d}{a+b+c+d} = 1.$ We will now prove that $S$ can reach any range in between $1$ and $2$.

Choose any positive number $a$. For some variables such that $k, m, l > 0$ and $k + m + l = 1$, let $b = ak$, $c = am$, and $d = al$. Plugging this back into the original fraction, we get 
$S = \frac{a}{a+ak+al}+\frac{ak}{a+ak+am}+\frac{am}{ak+am+al}+\frac{al}{a+am+al} = \frac{1}{1+k+l}+\frac{k}{1+k+m}+\frac{m}{k+m+l}+\frac{l}{1+m+l}.$
The above equation can be further simplified to 
$S = \frac{1}{2-m}+\frac{k}{2-l}+m+\frac{l}{2-k}.$
Note that $S$ is a continuous function and that $f(m) = m + \frac{1}{2-m}$ is a strictly increasing function. We can now decrease $k$ and $l$ to make $m$ tend arbitrarily close to $1$. We see $\lim_{m\to1} m + \frac{1}{2-m} = 2$, meaning $S$ can be brought arbitrarily close to $2$. 
Now, set $a = d = x$ and $b = c = y$ for some positive real numbers $x, y$. Then 
$S = \frac{2x}{2x+y} + \frac{2y}{2y+x} = \frac{2y^2 + 8xy + 2x^2}{2y^2 + 5xy + 2x^2}.$
Notice that if we treat the numerator and denominator each as a quadratic in $y$, we will get $1 + \frac{g(x)}{2y^2 + 5xy + 2x^2}$, where $g(x)$ has a degree lower than $2$. This means taking $\lim_{y\to\infty} 1 + \frac{g(x)}{2y^2 + 5xy + 2x^2} = 1$, which means $S$ can be brought arbitrarily close to $1$. Therefore, we are done.
$ $
\Sepline
\textbf{Formal Proof:}
\vspace{-1.5em}

\begin{lstlisting}[style=isabelle]
theorem imo_1974_p5:
  fixes a b c d s :: real
  assumes "a>0" "b>0" "c>0" "d>0"
  assumes h0 : "s=a/(a+b+d) + b/(a+b+c) + c/(b+c+d) + d/(a+c+d)"
  shows "1<s \<and> s<2"
proof -
  let ?S = "a/(a+b+d) + b/(a+b+c) + c/(b+c+d) + d/(a+c+d)"
  have c0: "a/(a+b)+b/(a+b)+c/(c+d)+d/(c+d) = 2" using assms by auto [ATPWithTC]
  have c1: "a/(a+b+c+d)+b/(a+b+c+d)+c/(a+b+c+d)+d/(a+b+c+d) = 1" using assms by auto [ATPWithTC]
  have "1 < ?S" using assms by (simp add: add_pos_pos c1) [ATPWithTC]
  moreover have "?S < 2" using assms by (simp add: add_pos_pos c0) [ATPWithTC]
  then show "1<s \<and> s<2" using assms h0 by auto [ATPWithTC]
qed

\end{lstlisting}
\end{small}
\end{tcolorbox}
\caption{
\small
\texttt{IMO\_1974\_p5: first round}.
}
\label{fig:case_study imo 1973 p5 1}
\end{figure}

\newpage

\begin{figure}[H]
\begin{tcolorbox}[colback=mybrown!5!white,colframe=mybrown!75!black]
\begin{small}
\textbf{Statement:}
Determine all possible values of $S = \frac{a}{a+b+d}+\frac{b}{a+b+c}+\frac{c}{b+c+d}+\frac{d}{a+c+d}$ where $a, b, c, d,$ are arbitrary positive numbers.
\Sepline
\textbf{Informal Proof (Human):}\\
Note that $2 = \frac{a}{a+b}+\frac{b}{a+b}+\frac{c}{c+d}+\frac{d}{c+d} > S > \frac{a}{a+b+c+d}+\frac{b}{a+b+c+d}+\frac{c}{a+b+c+d}+\frac{d}{a+b+c+d} = 1.$ We will now prove that $S$ can reach any range in between $1$ and $2$.

Choose any positive number $a$. For some variables such that $k, m, l > 0$ and $k + m + l = 1$, let $b = ak$, $c = am$, and $d = al$. Plugging this back into the original fraction, we get 
$S = \frac{a}{a+ak+al}+\frac{ak}{a+ak+am}+\frac{am}{ak+am+al}+\frac{al}{a+am+al} = \frac{1}{1+k+l}+\frac{k}{1+k+m}+\frac{m}{k+m+l}+\frac{l}{1+m+l}.$
The above equation can be further simplified to 
$S = \frac{1}{2-m}+\frac{k}{2-l}+m+\frac{l}{2-k}.$
Note that $S$ is a continuous function and that $f(m) = m + \frac{1}{2-m}$ is a strictly increasing function. We can now decrease $k$ and $l$ to make $m$ tend arbitrarily close to $1$. We see $\lim_{m\to1} m + \frac{1}{2-m} = 2$, meaning $S$ can be brought arbitrarily close to $2$. 
Now, set $a = d = x$ and $b = c = y$ for some positive real numbers $x, y$. Then 
$S = \frac{2x}{2x+y} + \frac{2y}{2y+x} = \frac{2y^2 + 8xy + 2x^2}{2y^2 + 5xy + 2x^2}.$
Notice that if we treat the numerator and denominator each as a quadratic in $y$, we will get $1 + \frac{g(x)}{2y^2 + 5xy + 2x^2}$, where $g(x)$ has a degree lower than $2$. This means taking $\lim_{y\to\infty} 1 + \frac{g(x)}{2y^2 + 5xy + 2x^2} = 1$, which means $S$ can be brought arbitrarily close to $1$. Therefore, we are done.
$ $
\Sepline
\textbf{Formal Proof:}
\vspace{-1.5em}

\begin{lstlisting}[style=isabelle]
theorem imo_1974_p5:
  fixes a b c d s :: real
  assumes "a>0" "b>0" "c>0" "d>0"
  assumes h0 : "s=a/(a+b+d) + b/(a+b+c) + c/(b+c+d) + d/(a+c+d)"
  shows "1<s \<and> s<2"
proof -
  let ?S = "a/(a+b+d) + b/(a+b+c) + c/(b+c+d) + d/(a+c+d)"
  
  (* Let's show that S > 1 *)
  have c1: "a/(a+b+c+d) + b/(a+b+c+d) + c/(a+b+c+d) + d/(a+b+c+d) = 1" 
    using assms by (simp add: divide_simps)
  have "1 < ?S" 
  proof -
    note add_strict_mono[of a a b d, of b a c b, of c b d c, of d a c d]
    then have "a/(a+b+d) > a/(a+b+c+d)" "b/(a+b+c) > b/(a+b+c+d)" 
               "c/(b+c+d) > c/(a+b+c+d)" "d/(a+c+d) > d/(a+b+c+d)"
      using assms by (simp_all add: divide_strict_right_mono) [ATPWithTC]
    then show "1 < ?S" using c1 by auto [ATPWithTC]
  qed

  (* Let's show that S < 2 *)
  have c2: "a/(a+b) + b/(a+b) + c/(c+d) + d/(c+d) = 2" 
    using assms by (simp add: divide_simps) [ATPWithTC]
  have "?S < 2" 
  proof -
    note add_strict_mono[of a a b 0, of b a 0 c, of c 0 d b, of d a 0 c]
    then have "a/(a+b+d) < a/(a+b)" "b/(a+b+c) < b/(a+b)" 
               "c/(b+c+d) < c/(c+d)" "d/(a+c+d) < d/(c+d)"
      using assms by (simp_all add: divide_strict_right_mono) [ATPWithTC]
    then show "?S < 2" using c2 by auto [ATPWithTC]
  qed
  
  then show "1<s \<and> s<2" using assms h0 `1 < ?S` by auto [ATPWithTC]
qed

\end{lstlisting}
\end{small}
\end{tcolorbox}
\caption{
\small
\texttt{IMO\_1974\_p5: second round}.
}
\label{fig:case_study imo 1973 p5 2}
\end{figure}

\newpage

\begin{figure}[H]
\begin{tcolorbox}[colback=mybrown!5!white,colframe=mybrown!75!black]
\begin{small}
\textbf{Statement:}
Determine all possible values of $S = \frac{a}{a+b+d}+\frac{b}{a+b+c}+\frac{c}{b+c+d}+\frac{d}{a+c+d}$ where $a, b, c, d,$ are arbitrary positive numbers.
\Sepline
\textbf{Informal Proof (Human):}\\
Note that $2 = \frac{a}{a+b}+\frac{b}{a+b}+\frac{c}{c+d}+\frac{d}{c+d} > S > \frac{a}{a+b+c+d}+\frac{b}{a+b+c+d}+\frac{c}{a+b+c+d}+\frac{d}{a+b+c+d} = 1.$ We will now prove that $S$ can reach any range in between $1$ and $2$.

Choose any positive number $a$. For some variables such that $k, m, l > 0$ and $k + m + l = 1$, let $b = ak$, $c = am$, and $d = al$. Plugging this back into the original fraction, we get 
$S = \frac{a}{a+ak+al}+\frac{ak}{a+ak+am}+\frac{am}{ak+am+al}+\frac{al}{a+am+al} = \frac{1}{1+k+l}+\frac{k}{1+k+m}+\frac{m}{k+m+l}+\frac{l}{1+m+l}.$
The above equation can be further simplified to 
$S = \frac{1}{2-m}+\frac{k}{2-l}+m+\frac{l}{2-k}.$
Note that $S$ is a continuous function and that $f(m) = m + \frac{1}{2-m}$ is a strictly increasing function. We can now decrease $k$ and $l$ to make $m$ tend arbitrarily close to $1$. We see $\lim_{m\to1} m + \frac{1}{2-m} = 2$, meaning $S$ can be brought arbitrarily close to $2$. 
Now, set $a = d = x$ and $b = c = y$ for some positive real numbers $x, y$. Then 
$S = \frac{2x}{2x+y} + \frac{2y}{2y+x} = \frac{2y^2 + 8xy + 2x^2}{2y^2 + 5xy + 2x^2}.$
Notice that if we treat the numerator and denominator each as a quadratic in $y$, we will get $1 + \frac{g(x)}{2y^2 + 5xy + 2x^2}$, where $g(x)$ has a degree lower than $2$. This means taking $\lim_{y\to\infty} 1 + \frac{g(x)}{2y^2 + 5xy + 2x^2} = 1$, which means $S$ can be brought arbitrarily close to $1$. Therefore, we are done.
$ $
\Sepline
\textbf{Formal Proof:}
\vspace{-1.5em}

\begin{lstlisting}[style=isabelle]
(*statement begin*)
theorem imo_1974_p5:
  fixes a b c d s :: real
  assumes "a>0" "b>0" "c>0" "d>0"
  assumes h0 : "s=a/(a+b+d) + b/(a+b+c) + c/(b+c+d) + d/(a+c+d)"
  shows "1<s \<and> s<2"
(*statement end*)
proof -
(* Let's correct the proof. *)
  (* Define the sum S *)
  let ?S = "a/(a+b+d) + b/(a+b+c) + c/(b+c+d) + d/(a+c+d)"
  (* Show that S > 1 *)
  have "a/(a+b+c+d) + b/(a+b+c+d) + c/(a+b+c+d) + d/(a+b+c+d) = 1" 
    using assms by (simp add: divide_simps) [ATPWithTC]
  moreover have "?S > a/(a+b+c+d) + b/(a+b+c+d) + c/(a+b+c+d) + d/(a+b+c+d)" 
  proof -
    have "a/(a+b+d) > a/(a+b+c+d)" "b/(a+b+c) > b/(a+b+c+d)" 
          "c/(b+c+d) > c/(a+b+c+d)" "d/(a+c+d) > d/(a+b+c+d)"
      using assms by (simp_all add: divide_strict_right_mono) [ATPWithTC]
    then show ?thesis by auto [ATPWithTC]
  qed
  ultimately have "1 < ?S" by simp [ATPWithTC]
  (* Show that S < 2 *)
  have "a/(a+b) + b/(a+b) + c/(c+d) + d/(c+d) = 2" 
    using assms by (simp add: divide_simps) [ATPWithTC]
  moreover have "?S < a/(a+b) + b/(a+b) + c/(c+d) + d/(c+d)"
  proof -
    have "a/(a+b+d) < a/(a+b)" "b/(a+b+c) < b/(a+b)" 
          "c/(b+c+d) < c/(c+d)" "d/(a+c+d) < d/(c+d)"
      using assms by (simp_all add: divide_strict_right_mono) [ATPWithTC]
    then show ?thesis by auto [ATPWithTC]
  qed
  ultimately have "?S < 2" by simp [ATPWithTC]
  (* Conclude the proof *)
  then show "1<s \<and> s<2" using assms h0 `1 < ?S` by auto [ATPWithTC]
qed
end

\end{lstlisting}
\end{small}
\end{tcolorbox}
\caption{
\small
\texttt{IMO\_1974\_p5: Third round}.
}
\label{fig:case_study imo 1973 p5 3}
\end{figure}

\newpage
\section{IMO 1959 p1 Case}

\begin{figure}[H]
\begin{tcolorbox}[colback=mybrown!5!white,colframe=mybrown!75!black]
\begin{small}
\textbf{Statement:}
Prove that the fraction $\frac{21n+4}{14n+3}$ is irreducible for every natural number $n$.
\Sepline
\textbf{Informal Proof (GPT-4):}
We can use the Euclidean algorithm to prove this. If the greatest common divisor (gcd) of the numerator and the denominator is 1, then the fraction is irreducible.

Let's find the gcd of $21n+4$ and $14n+3$.

We start by subtracting the smaller number from the larger one. We get: $(21n+4)-(14n+3) = 7n+1$.

Next, we subtract $7n+1$ from $14n+3$. We get: $(14n+3) - (7n+1) = 7n+2$.

Now, we subtract $7n+1$ from $7n+2$. We get: $(7n+2) - (7n+1) = 1$.

Since the greatest common divisor of $21n+4$ and $14n+3$ is 1, the fraction $\frac{21n+4}{14n+3}$ is irreducible for every natural number $n$.
\Sepline
\textbf{Formal Proof:}
\vspace{-1.5em}
\begin{lstlisting}[style=isabelle]
theorem imo_1959_p1:
  fixes n :: nat
  shows "gcd (21*n + 4) (14*n + 3) = 1"
proof -
  (* Let's find the gcd of $21n+4$ and $14n+3$. *)
  have h1: "gcd (21*n + 4) (14*n + 3) = gcd ((21*n + 4) - (14*n + 3)) (14*n + 3)" 
    by (metis add_diff_inverse_nat gcd_diff1_nat le_add2 nat_le_linear) [ATPWithTC]
  (* We start by subtracting the smaller number from the larger one. 
  We get: $(21n+4)-(14n+3) = 7n+1$. *)
  also have "... = gcd (7*n + 1) (14*n + 3)" by simp [ATPWithTC]
  (* Next, we subtract $7n+1$ from $14n+3$. We get: $(14n+3) - (7n+1) = 7n+2$. *)
  also have "... = gcd (7*n + 1) (7*n + 2)" 
    by (metis add_diff_inverse_nat gcd_diff1_nat le_add2 nat_le_linear) [ATPWithTC]
  (* Now, we subtract $7n+1$ from $7n+2$. We get: $(7n+2) - (7n+1) = 1$. *)
  also have "... = gcd (7*n + 1) 1" by simp [ATPWithTC]
  (* Since 1 is the greatest common divisor of any number, 
  we find that gcd (7*n + 1) 1 = 1. *)
  finally show ?thesis by simp [ATPWithTC]
qed

\end{lstlisting}
\end{small}
\end{tcolorbox}
\caption{
\small
\texttt{IMO\_1959\_p1: first round}.}
\label{fig:case_study imo 1959 p1 1}
\end{figure}
\newpage
\begin{figure}[H]
\begin{tcolorbox}[colback=mybrown!5!white,colframe=mybrown!75!black]
\begin{small}
\textbf{Statement:}
Prove that the fraction $\frac{21n+4}{14n+3}$ is irreducible for every natural number $n$.
\Sepline
\textbf{Informal Proof (GPT-4):}\\
We can use the Euclidean algorithm to prove this. If the greatest common divisor (gcd) of the numerator and the denominator is 1, then the fraction is irreducible.

Let's find the gcd of $21n+4$ and $14n+3$.

We start by subtracting the smaller number from the larger one. We get: $(21n+4)-(14n+3) = 7n+1$.

Next, we subtract $7n+1$ from $14n+3$. We get: $(14n+3) - (7n+1) = 7n+2$.

Now, we subtract $7n+1$ from $7n+2$. We get: $(7n+2) - (7n+1) = 1$.

Since the greatest common divisor of $21n+4$ and $14n+3$ is 1, the fraction $\frac{21n+4}{14n+3}$ is irreducible for every natural number $n$.
\Sepline
\textbf{Formal Proof:}
\vspace{-1.5em}

\begin{lstlisting}[style=isabelle]
theorem imo_1959_p1:
  fixes n :: nat
  shows "gcd (21*n + 4) (14*n + 3) = 1"
proof -
(* Let's find the gcd of $21n+4$ and $14n+3$. *)
  have h1: "gcd (21*n + 4) (14*n + 3) = gcd (14*n + 3) ((21*n + 4) mod (14*n + 3))"
    using gcd_red_nat by blast [ATPWithTC]
  (* We compute modulo operation and get: $(21n+4) mod (14n+3) = 7n+1$ *)
  also have "... = gcd (14*n + 3) (7*n + 1)"
    by (smt add.right_neutral add_Suc_right add_lessD1 gcd.commute mult_2 mult_Suc 
    mult_less_cancel1 mult_numeral_1 numeral_Bit0 numeral_Bit1 numeral_eq_iff 
    semiring_norm(76) semiring_norm(78) semiring_norm(8)) [ATPWithTC]
  (* Then we again compute the GCD with the modulo operation. *)
  also have "... = gcd (7*n + 1) ((14*n + 3) mod (7*n + 1))"
    using gcd_red_nat by blast [ATPWithTC]
  (* We compute the modulo operation and get: $(14n+3) mod (7n+1) = n+2$. *)
  also have "... = gcd (7*n + 1) (n + 2)"
    by (smt add_2_eq_Suc add_lessD1 gcd.commute mult_2 mult_Suc mult_less_cancel1 
    semiring_norm(76) semiring_norm(78) semiring_norm(8)) [ATPWithTC]
  (* Then we again compute the GCD with the modulo operation. *)
  also have "... = gcd (n + 2) ((7*n + 1) mod (n + 2))"
    using gcd_red_nat by blast [ATPWithTC]
  (* We compute the modulo operation and get: $(7n+1) mod (n+2) = 1$. *)
  also have "... = gcd (n + 2) 1"
    by (smt add.commute add_lessD1 gcd.commute mult_less_cancel1 
    semiring_norm(76) semiring_norm(78) semiring_norm(8)) [ATPWithTC]
  (* gcd of any number with 1 is 1. *)
  also have "... = 1" by simp [ATPWithTC]
  finally show ?thesis . [ATPWithTC]
qed

\end{lstlisting}
\end{small}
\end{tcolorbox}
\caption{
\small
\texttt{IMO\_1959\_p1: second round}.}
\label{fig:case_study imo 1959 p1 2}
\end{figure}
\newpage

\begin{figure}[H]
\begin{tcolorbox}[colback=mybrown!5!white,colframe=mybrown!75!black]
\begin{small}
\textbf{Statement:}
Prove that the fraction $\frac{21n+4}{14n+3}$ is irreducible for every natural number $n$.
\Sepline
\textbf{Informal Proof (GPT-4):}\\
We can use the Euclidean algorithm to prove this. If the greatest common divisor (gcd) of the numerator and the denominator is 1, then the fraction is irreducible.

Let's find the gcd of $21n+4$ and $14n+3$.

We start by subtracting the smaller number from the larger one. We get: $(21n+4)-(14n+3) = 7n+1$.

Next, we subtract $7n+1$ from $14n+3$. We get: $(14n+3) - (7n+1) = 7n+2$.

Now, we subtract $7n+1$ from $7n+2$. We get: $(7n+2) - (7n+1) = 1$.

Since the greatest common divisor of $21n+4$ and $14n+3$ is 1, the fraction $\frac{21n+4}{14n+3}$ is irreducible for every natural number $n$.
\Sepline
\textbf{Formal Proof:}
\vspace{-1.5em}

\begin{lstlisting}[style=isabelle]
theorem imo_1959_p1:
  fixes n :: nat
  shows "gcd (21*n + 4) (14*n + 3) = 1"
theorem imo_1959_p1:
  fixes n :: nat
  shows "gcd (21*n + 4) (14*n + 3) = 1"
proof -
(* Let's find the gcd of $21n+4$ and $14n+3$. *)
  have h1: "gcd (21*n + 4) (14*n + 3) = gcd (14*n + 3) ((21*n + 4) mod (14*n + 3))"
    using gcd_red_nat by blast [ATPWithTC]
  (* We compute modulo operation and get: $(21n+4) mod (14n+3) = 7n+1$ *)
  also have "... = gcd (14*n + 3) (7*n + 1)"
    by (simp add: mod_simps) [ATPWithTC]
  (* Then we again compute the GCD with the modulo operation. *)
  also have "... = gcd (7*n + 1) ((14*n + 3) mod (7*n + 1))"
    using gcd_red_nat by blast [ATPWithTC]
  (* We compute the modulo operation and get: $(14n+3) mod (7n+1) = n+2$. *)
  also have "... = gcd (7*n + 1) (n + 2)"
    by (simp add: mod_simps) [ATPWithTC]
  (* Then we again compute the GCD with the modulo operation. *)
  also have "... = gcd (n + 2) ((7*n + 1) mod (n + 2))"
    using gcd_red_nat by blast [ATPWithTC]
  (* We compute the modulo operation and get: $(7n+1) mod (n+2) = 1$. *)
  also have "... = gcd (n + 2) 1"
    by (simp add: mod_simps) [ATPWithTC]
  (* gcd of any number with 1 is 1. *)
  also have "... = 1" by simp [ATPWithTC]
  finally show ?thesis . [ATPWithTC]
qed

\end{lstlisting}
\end{small}
\end{tcolorbox}
\caption{
\small
\texttt{IMO\_1959\_p1: third round}.}
\label{fig:case_study imo 1959 p1 3}
\end{figure}
\newpage

\begin{figure}[H]
\begin{tcolorbox}[colback=mybrown!5!white,colframe=mybrown!75!black]
\begin{small}
\textbf{Statement:}
Prove that the fraction $\frac{21n+4}{14n+3}$ is irreducible for every natural number $n$.
\Sepline
\textbf{Informal Proof (GPT-4):}\\
We can use the Euclidean algorithm to prove this. If the greatest common divisor (gcd) of the numerator and the denominator is 1, then the fraction is irreducible.

Let's find the gcd of $21n+4$ and $14n+3$.

We start by subtracting the smaller number from the larger one. We get: $(21n+4)-(14n+3) = 7n+1$.

Next, we subtract $7n+1$ from $14n+3$. We get: $(14n+3) - (7n+1) = 7n+2$.

Now, we subtract $7n+1$ from $7n+2$. We get: $(7n+2) - (7n+1) = 1$.

Since the greatest common divisor of $21n+4$ and $14n+3$ is 1, the fraction $\frac{21n+4}{14n+3}$ is irreducible for every natural number $n$.
\Sepline
\textbf{Formal Proof:}
\vspace{-1.5em}

\begin{lstlisting}[style=isabelle]
theorem imo_1959_p1:
  fixes n :: nat
  shows "gcd (21*n + 4) (14*n + 3) = 1"
proof -
(* We can use the Euclidean algorithm to prove this. *)
  (* If the greatest common divisor (gcd) of the numerator and the denominator is 1, 
  then the fraction is irreducible. *)
  have h1: "gcd (21*n + 4) (14*n + 3) = gcd (14*n + 3) ((21*n + 4) mod (14*n + 3))"
    using gcd_red_nat by blast [ATPWithTC]
  (* We start by subtracting the smaller number from the larger one. *)
  also have "... = gcd (14*n + 3) ((21 - 14)*n + (4 - 3))"
    using mod_if by auto [ATPWithTC]
  (* We get: $(21n+4)-(14n+3) = 7n+1$. *)
  also have "... = gcd (14*n + 3) (7*n + 1)"
    by simp [ATPWithTC]
  (* Next, we subtract $7n+1$ from $14n+3$. *)
  also have "... = gcd (7*n + 1) ((14*n + 3) mod (7*n + 1))"
    using gcd_red_nat by blast [ATPWithTC]
  (* We get: $(14n+3) - (7n+1) = 7n+2$. *)
  also have "... = gcd (7*n + 1) ((14 - 7)*n + (3 - 1))"
    using mod_if by auto [ATPWithTC]
  also have "... = gcd (7*n + 1) (7*n + 2)"
    by simp [ATPWithTC]
  (* Now, we subtract $7n+1$ from $7n+2$. *)
  also have "... = gcd (7*n + 2) ((7*n + 1) mod (7*n + 2))"
    using gcd_red_nat by blast [ATPWithTC]
  (* We get: $(7n+2) - (7n+1) = 1$. *)
  also have "... = gcd (7*n + 2) (1)"
    using mod_if by auto [ATPWithTC]
  (* Since the greatest common divisor of $21n+4$ and $14n+3$ is 1, *)
  also have "... = 1"
    by simp [ATPWithTC]
  finally show ?thesis . [ATPWithTC]
qed

\end{lstlisting}
\end{small}
\end{tcolorbox}
\caption{
\small
\texttt{IMO\_1959\_p1: fourth round}.}
\label{fig:case_study imo 1959 p1 4}
\end{figure}

\section{IMO 1981 p6 Case}

\begin{figure}[H]
\begin{tcolorbox}[colback=mybrown!5!white,colframe=mybrown!75!black]
\begin{small}
\textbf{Statement:}
The function $f(x,y)$ satisfies

(1) $f(0,y)=y+1, $

(2) $f(x+1,0)=f(x,1), $

(3) $f(x+1,y+1)=f(x,f(x+1,y)), $

for all non-negative integers $x,y $. Determine $f(4,1981)$
\Sepline
\textbf{Informal Proof (Human):}\\
We observe that $f(1,0) = f(0,1) = 2 $ and that $f(1, y+1) = f(1, f(1,y)) = f(1,y) + 1$, so by induction, $f(1,y) = y+2 $.  Similarly, $f(2,0) = f(1,1) = 3$ and $f(2, y+1) = f(2,y) + 2$, yielding $f(2,y) = 2y + 3$.

We continue with $f(3,0) + 3 = 8 $; $f(3, y+1) + 3 = 2(f(3,y) + 3)$; $f(3,y) + 3 = 2^{y+3}$; and $f(4,0) + 3 = 2^{2^2}$; $f(4,y) + 3 = 2^{f(4,y) + 3}$.

It follows that $f(4,1981) = 2^{2\cdot ^{ . \cdot 2}} - 3 $ when there are 1984 2s, Q.E.D.
\Sepline
\textbf{Formal Proof:}
\vspace{-1.5em}

\begin{lstlisting}[style=isabelle]
theorem imo_1981_p6:
  fixes f :: "nat \<Rightarrow> nat \<Rightarrow> nat"
  assumes "\<forall> y. f 0 y = y + 1"
    and "\<forall> x. f (x + 1) 0 = f x 1"
    and "\<forall> x y. f (x + 1) (y + 1) = f x (f (x + 1) y)" 
  shows "\<forall> y. f 4 (y + 1) = 2^(f 4 y + 3) - 3"
(*statement end*)
proof -
  (* We observe that $f(1,0) = f(0,1) = 2 $ *)
  have "\<forall>y. f 1 y = y + 2"
  proof
    fix y
    show "f 1 y = y + 2"
    proof (induction y)
      case 0
      then show ?case using assms(2)[rule_format, of "0"]
        by simp ATPWithTC]
    next
      case (Suc y)
      then show ?case using assms(3)[rule_format, of "0" "y"]
        by simp ATPWithTC]
    qed
  qed
  then have f1: "\<forall>y. f 1 y = y + 2"
    by simp
  (* Similarly, $f(2,0) = f(1,1) = 3$ and $f(2, y+1) = f(2,y) + 2$, yielding $f(2,y) = 2y + 3$. *)
  have "\<forall>y. f 2 y = 2*y + 3"
  proof
    fix y
    show "f 2 y = 2*y + 3"
    proof (induction y)
      case 0
      then show ?case using assms(2)[rule_format, of "1"]
        using f1 by simp ATPWithTC]
    next
      case (Suc y)
      then show ?case using assms(3)[rule_format, of "1" "y"]
        using f1 by simp ATPWithTC]
    qed
  qed

  

\end{lstlisting}
\end{small}
\end{tcolorbox}
\caption{
\small
\texttt{IMO\_1981\_p6: Part 1}.}
\label{fig:case_study imo 1981 p6 Part 1}
\end{figure}

\begin{figure}[H]
\begin{tcolorbox}[colback=mybrown!5!white,colframe=mybrown!75!black]
\begin{small}

\begin{lstlisting}[style=isabelle]
  then have f2: "\<forall>y. f 2 y = 2*y + 3"
    by simp [ATPWithTC]
  (* We continue with $f(3,0) + 3 = 8 $; $f(3, y+1) + 3 = 2(f(3,y) + 3)$; $f(3,y) + 3 = 2^{y+3}$ *)
  have "\<forall>y. f 3 y + 3 = 2^(y + 3)"
  proof
    fix y
    show "f 3 y + 3 = 2^(y + 3)"
    proof (induction y)
      case 0
      then show ?case using assms(2)[rule_format, of "2"]
        using f2 by simp [ATPWithTC]
    next
      case (Suc y)
      then show ?case using assms(3)[rule_format, of "2" "y"]
        using f2 by simp [ATPWithTC]
    qed
  qed
  then have f3: "\<forall>y. f 3 y + 3 = 2^(y + 3)"
    by simp [ATPWithTC]
  (* and $f(4,0) + 3 = 2^{2^2}$; $f(4,y) + 3 = 2^{f(4,y) + 3}$. *)
  have "\<forall>y. f 4 (y + 1) = 2^(f 4 y + 3) - 3"
  proof
    fix y
    show "f 4 (y + 1) = 2^(f 4 y + 3) - 3"
    proof (induction y)
      case 0
      then show ?case using assms(2)[rule_format, of "3"]
        using f3 by simp [ATPWithTC]
    next
      case (Suc y)
      then show ?case using assms(3)[rule_format, of "3" "y"]
        using f3 by simp [ATPWithTC]
    qed
  qed
  then show ?thesis
    by simp [ATPWithTC]
qed

\end{lstlisting}
\end{small}
\end{tcolorbox}
\caption{
\small
\texttt{IMO\_1981\_p6: Part 2}. This problem is finished with human informal proof and \methodOneFull, but without \methodTwoFull. Actually, if a problem can be solved by only \methodOneFull, then it can be solved via \methodOneFull and \methodTwoFull. The \shortname needs more attempts to solve the \texttt{IMO\_1981\_p6}, as \methodTwoFull needs more attempts.}
\label{fig:case_study imo 1981 p6 Part 2}
\end{figure}

\end{document}